\documentclass[runningheads]{llncs}
\usepackage{cite}
\usepackage{amsmath,amssymb,amsfonts}
\usepackage{algorithmic}
\usepackage{textcomp}
\usepackage{xcolor}
\usepackage{comment}
\usepackage{multirow}
\usepackage{hyperref}  
\usepackage{cleveref}

\usepackage[T1]{fontenc}
\usepackage{authblk}
\usepackage{graphicx}
\begin{document}
\title{Celeb-FBI: A Benchmark Dataset on Human Full Body Images and Age, Gender, Height and Weight Estimation using Deep Learning Approach}
%
%
\author{Pronay Debnath\inst{1} \and
Usafa Akther Rifa\inst{2} \and
Busra Kamal Rafa\inst{3} \and
Ali Haider Talukder Akib\inst{4} \and
Md. Aminur Rahman\inst{5}}

\institute{ {Department of Computer Science and Engineering,\\
Ahsanullah University of Science and Technology, Dhaka, Bangladesh} \\
\email{\{pronaydebnath99 \inst{1},
usafarifa97\inst{2},
brafa263.3\inst{3},
alihaiderakib\inst{4},
aminur.rahman.rsd\inst{5}\}}@gmail.com}

\authorrunning{Pronay et al.}
%
%
\maketitle              
\begin{abstract}
The scarcity of comprehensive datasets in surveillance, identification, image retrieval systems, and healthcare poses a significant challenge for researchers in exploring new methodologies and advancing knowledge in these respective fields. Furthermore, the need for full-body image datasets with detailed attributes like height, weight, age, and gender is particularly significant in areas such as fashion industry analytics, ergonomic design assessment, virtual reality avatar creation, and sports performance analysis. To address this gap, we have created the ‘Celeb-FBI' dataset which contains 7,211 full-body images of individuals accompanied by detailed information on their height, age, weight, and gender. Following the dataset creation, we proceed with the preprocessing stages, including image cleaning, scaling, and the application of Synthetic Minority Oversampling Technique (SMOTE). Subsequently, utilizing this prepared dataset, we employed three deep learning approaches: Convolutional Neural Network (CNN), 50-layer ResNet, and 16-layer VGG, which are used for estimating height, weight, age, and gender from human full-body images. From the results obtained, ResNet-50 performed best for the system with an accuracy rate of 79.18\% for age, 95.43\% for gender, 85.60\% for height and 81.91\% for weight.


\end{abstract}

\section{Introduction}
This chapter introduces a new sophisticated benchmark task, that entails predicting an individual's gender along with additional biometric information (age, height, and weight) from their full body image within a predefined range as a multilabel classification problem. For this task, we compile the corresponding dataset for training and testing purposes. While there has been notable progress in the other type of face recognition task, particularly in face identification \cite{b1,b2,b3,b4}, this chapter aims to rectify current shortcomings within this domain, contributing to the ongoing advancement of computer vision \cite{b5}. First, in current biometrics research using people images, the main focus is on identifying and finding only similar images. Additionally, there is exploration into gender prediction, along with concurrent estimations of age, or the prediction of height, weight, and Body Mass Index (BMI) from facial images. Rather than limiting the focus to specific aspects, it is more practical in face detection and recognition methods to encompass all these biometric data from a single image which is missing. Second, about the scarcity of publicly available datasets containing facial images, let alone those encompassing full-body images. Extensive datasets featuring a diverse range of individuals play a vital role in face identification research implementation. Their accessibility is a significant challenge for many researchers, particularly those in academia, limiting valuable contributions.
Our benchmark task defines the following properties. One, we introduce a dataset that determines the biometric information of a person which is more natural to take place in practical and to many real life applications such as security authentication, content-based image retrieval, surveillance system, biometry etc. Two, this public dataset is based on the full body images of popular celebrities with their biometric information in a range. As it is important in image interpretation and recognition to have a variety of people for machine learning. And it could be directly utilized in a broad spectrum of real-world scenarios, subject to specific licensing agreements. Three, we summed up images of 7,211 celebrities from \url{www.google.com} and \url{www.bing.com} and provided their personal attributes, among them 4,377 are female and 2,834 are male. We consider a specific range for the personal attributes as humans do look different on the basis of their genetics even if with the same height, weight or age. To predict these personal attributes most accurately by just looking at their appearance is hard for even humans. Thus difficulty arises assuming attributes prediction as a regression problem. But predicting them within a defined range becomes more reasonable. So we assumed it as a classification problem and motivating researchers to develop recognizers capable of identifying individual entities in diverse domains, including medical and security applications, beyond conventional use. And it is made sure  there is no duplicate image present in the dataset.
\newline
This task tends to meet some following challenges. With the increase of class variance, such as for age exceeds 80, weight goes beyond 60 kg, and uncommon instances of height, there is a likelihood of reduced proper and standard image data due to difficulties in finding. These lead to an imbalanced dataset. However, to improve the performance of our task we applied smote to create artificial samples for all attributes on the training portion. In addition to introducing this challenging yet interesting benchmark task, we provide a training dataset to support the task. Manually selected from our list of 9,500 top celebrities, this training dataset consists of about 7,211 images. 
\newline 
To the best of our knowledge, our training data based on this task, is the first publicly available one till now. We plan to make the dataset even larger in the near future. Using this training data, we trained three distinct models employing a classification setup, treating each attribute's unique value as an individual class. The experimental outcomes indicate that the ResNet-50 model structure performs notably well compared to other models.

\section{Related Work}
Numerous literary works explore predicting height, weight, age, and gender from human facial or body images. Can Yilmaz Altinigne et al. \cite{b6} estimated height and weight from single, unconstrained images, without requiring prior information on individual characteristics such as age, gender, or facial features using the U-Net model \cite{b7}. The authors utilized frontal full body unconstrained images from the IMDB dataset with height values only, which was originally employed by Gunel \cite{b8}. Antitza Dantcheva INRIA et al. \cite{b9} predicted height, weight, and body mass index from facial images. This paper introduced a new way to predict height, weight, and BMI using ResNet-50 from just a single facial image. They worked on a novel dataset consisting of 1026 subjects collected from the WWW. Min Jiang et al. \cite{b10} examined the feasibility of body weight analysis using the visual cues found in photographs of human bodies. For the purpose of the study, a visual-body-to-BMI dataset comprising 5900 photos of 2950 people and the labels for their gender, height, and weight were gathered and cleaned in this research. Carmelo Velardo et al. \cite{b11} offered a technique for estimating a person's weight that makes use of anthropometric characteristics, which are known to be connected to a person's appearance and weight. They used the NHANES \cite{b12} dataset for their experiment. Dhanamjayulu C et al. \cite{b13} investigated the relationship between facial appearance and body weight, focusing on estimating Body Mass Index (BMI) values from human face images. In this research, the residual neural network model outperformed the others in terms of correlation values, area under the curve metrics, and mean absolute errors (MAEs). The dataset was gathered from the internet, cleansed, and applied to the problem of BMI estimation. Abdullah M. et al. \cite{b14} used the UP students’ dataset in their research. The dataset contained both gender categories (male and female). It included pictures of around 430 pupils, ages ranging from 20 to 55, including 233 males and 195 females. From this research, while age prediction accuracy was roughly 57\% for both genders combined, gender prediction attained an overall accuracy of about 82\%. Semih Gunel et al. \cite{b26} assembled the biggest height estimation dataset, which included 274,964 photos and 12,104 actors from the IMDB database. Their study came to the conclusion that persistent geometric scale ambiguity makes height estimation problematic, highlighting the significance of tackling this issue in tasks such as 3D human pose estimation. Koichi Ito et al. \cite{b15} used the IMDB-WIKI dataset \cite{b16}, which consists of 523,051 images. This research presented a CNN-based technique for predicting gender and age from face photos and investigated several architectures, techniques for regression and classification, and learning strategies for single- and multi-tasking.

\section{Dataset}
\subsection{Dataset Collection}
\begin{figure}[htbp]
    \centering
    \includegraphics[width=85mm,scale=0.4]{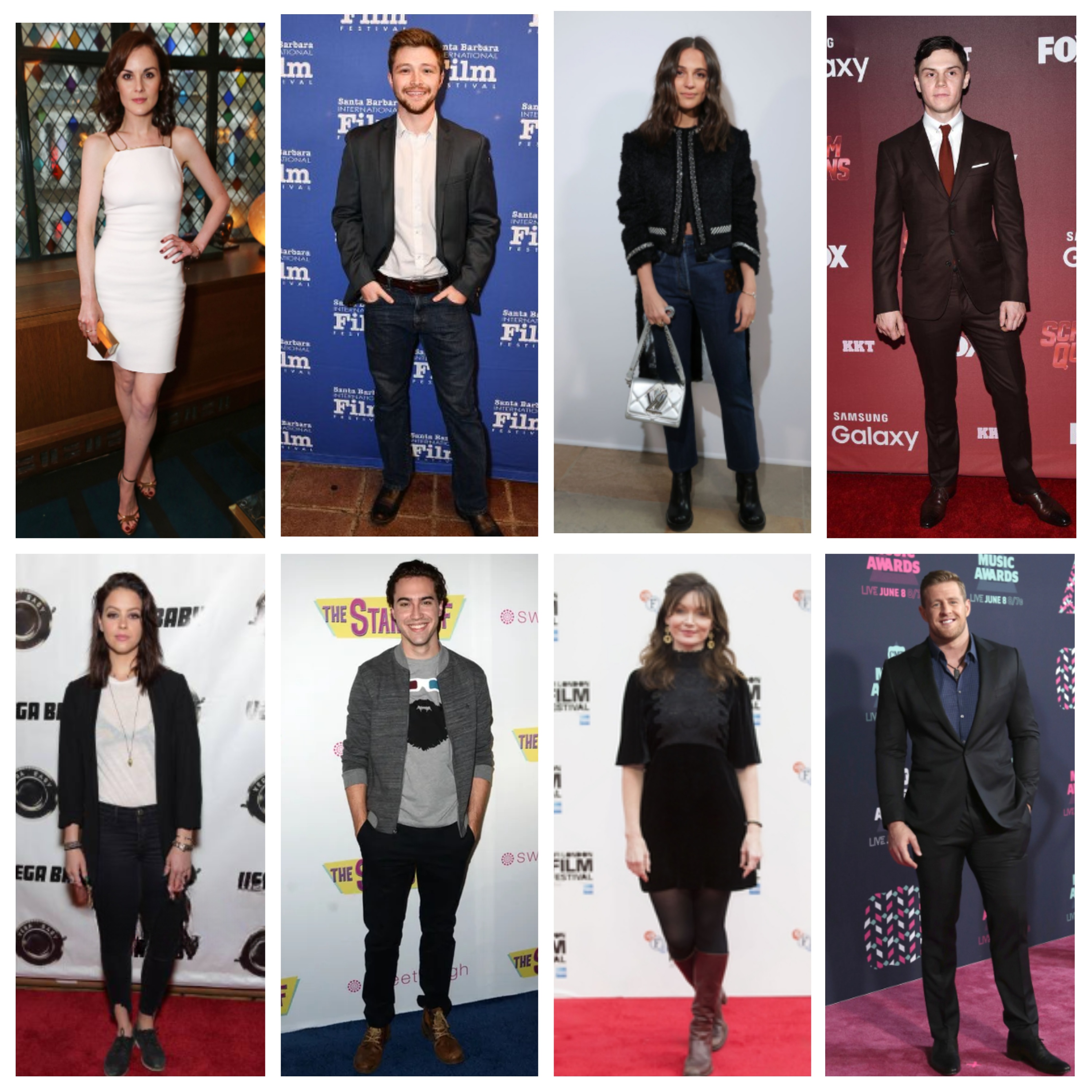}
    \caption{Sample of collected images}
    \label{fig:single_img}
\end{figure}

We assembled a dataset named ‘Celeb-FBI’ encompassing 9,500 subjects, of which 7,211 were accompanied by full-body images of different celebrities from various regions like America, Asia, Europe etc. Data that had no images online and data that had images but poor resolution quality were cleaned, resulting in the subjects being reduced from 9500 to 7211. These data were collated from tabulated sources such as \url{www.bodysize.org} and \url{www.howtallis.org}. These data were meticulously collected and stored in an Excel spreadsheet which were kept in columns such as serial number, name of celebrity, height, weight, gender, and age. We took front-facing and standing full-body images, as illustrated in Fig.~\ref{fig:single_img} which were sourced from \url{www.google.com}  and \url{www.bing.com} ensuring that they were not protected by copyright. All the images are in png format and renamed as \texttt{SerialNo\_(height)h\_(weight)w\_gender\_(age)a.png} where “h”, “w”, and “a” denote height, weight, and age respectively. This meticulous approach facilitates subsequent data analysis and the formulation of distinct classes. Utilizing our comprehensive dataset, we trained the deep learning models to proficiently estimate age, gender, height and weight. To obtain access to the dataset, one can submit a request to \href{mailto:pronaydebnath99@gmail.com}{pronaydebnath99@gmail.com}.

\subsection{Dataset validation}
The individuals responsible for data collection thoroughly validated every piece of information. This included information on height, weight, age and gender that was entered into Excel file along with related images. Validators carefully checked that all of the data, which they obtained from the internet and cross-referenced with images, was accurate. If a mismatch arose during the validation process, the expert reviewed the conflicting information and images to make a final decision. He likely relied on his expertise to assess the accuracy of the data and determine the most reliable source. Once resolved, the validated information was then incorporated into the Excel file for further analysis.

\subsection{Identity of validators}
We set standards in place to mitigate bias during the validation process. All the validators are undergraduate students aged between 22 and 24 years and an expert who is a 35 year old researcher specializing in NLP and computer vision.
\begin{table}[http]
\begin{center}
 \caption{Indentity of the validators}
  \label{tab:table2}
   
  \begin{tabular}{|p{1.8cm}|p{1.5cm}|p{1.5cm}|p{1.5cm}|p{1.5cm}|p{1.6cm}|}
   
     \hline
       \textbf{}&
        \textbf{\centering VLD-1}&
        \textbf{\centering VLD-2}&
        \textbf{\centering VLD-3}&
        \textbf{\centering VLD-4}&
        \textbf{\centering Expert}\\
    
        \hline
        \textbf{Research-Status} & Undergrad  & Undergrad & Undergrad & RA & NLP, CV researcher \\
        \hline
        \textbf{Research-field} & NLP, CV  & CV & NLP, HCI & NLP & NLP, CV, HCI \\
        \hline
        \textbf{Age} &  23  &  22 &  24 & 23 &  35 \\
        \hline
        \textbf{Gender} &  Male  &  Female &  Female &  Male &  Male \\
        \hline

    \end{tabular}
    \end{center}
\end{table}
\Cref{tab:table2} is an overview of the validators details, including their research status, field of research and personal attributes. Here VLD denotes validator here.

\subsection{Dataset Statistics}
Our dataset comprises full-body images accompanied by precise information on 7,211 subjects. However, upon applying specified ranges, the dataset was refined, resulting in reductions to 6,710 subjects for height, 7,139 subjects for age, and 5,941 subjects for weight. \Cref{tab:table3} and Fig.~\ref{fig:2} -- Fig.~\ref{fig:5} present all the information in an organised manner as shown below.

\begin{table}[http]
\begin{center}
 \caption{Dataset Statistics}
  \label{tab:table3}
   
  \begin{tabular}{|p{2.5cm}|p{1.2cm}|p{1.2cm}|p{1.8cm}|p{1.3cm}|p{2.5cm}|}
   
        \hline
        \textbf{}&
        \textbf{\centering Range}&
        \textbf{\centering Total classes}&
        \textbf{\centering Intra-class Gap}&
        \textbf{\centering No of Images}&
        \textbf{\centering No of Images After SMOTE}\\
    
        \hline
        \textbf{\centering Age (years)} & 21-80  & 12 & 5 years & 7139 & 16188 \\
        \hline
        \textbf{\centering Height\newline(feet and inch)} & 4.8-6.5 & 8 & 3 inch & 6710 & 14352 \\
        \hline
        \textbf{Weight (kg)} & 41-110  & 14 & 5 kg & 5941 & 15078 \\
        \hline
        \textbf{Gender} & -  & 2 & - & 7211 & 7888 \\
        \hline
        
    \end{tabular}
    \end{center}
\end{table}

\begin{figure*}[htbp] 
    \centering
    \begin{minipage}[b]{5.2cm} 
    \centering
    \includegraphics[width=\linewidth, height=5.2cm]{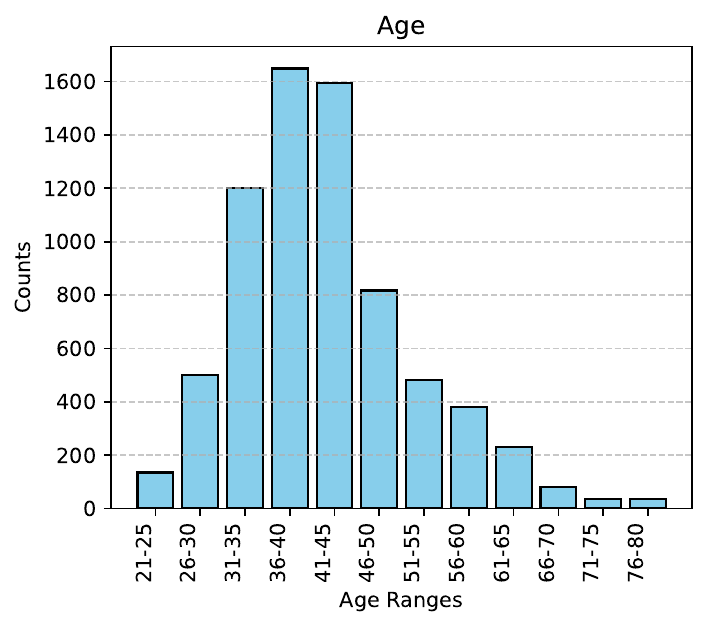}
    \caption{Dataset (Age)}
    \label{fig:2}
\end{minipage}
    \hfill 
    \begin{minipage}[b]{5.2cm} 
        \centering
        \includegraphics[width=\linewidth, height=5.2cm]{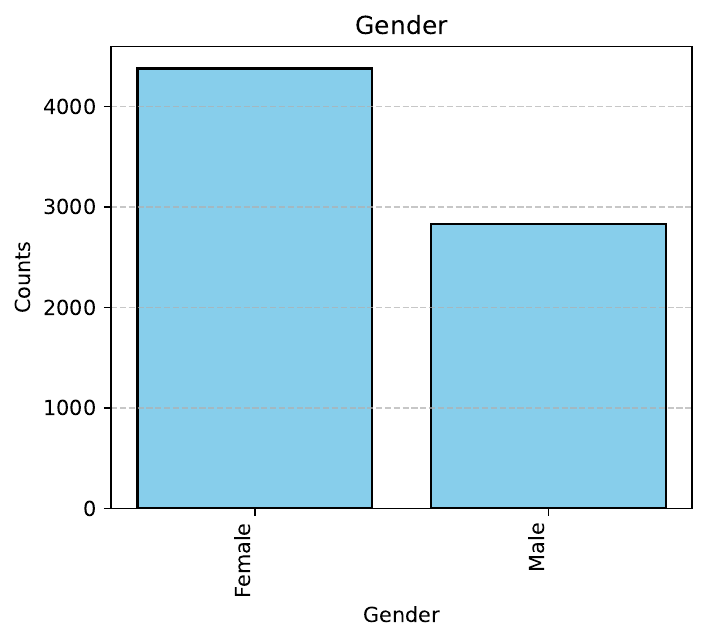}
        \caption{Dataset (Gender)}
        \label{fig:3}
    \end{minipage}
    \hfill 
    \begin{minipage}[b]{5.2cm} 
        \centering
        \includegraphics[width=\linewidth, height=5.2cm]{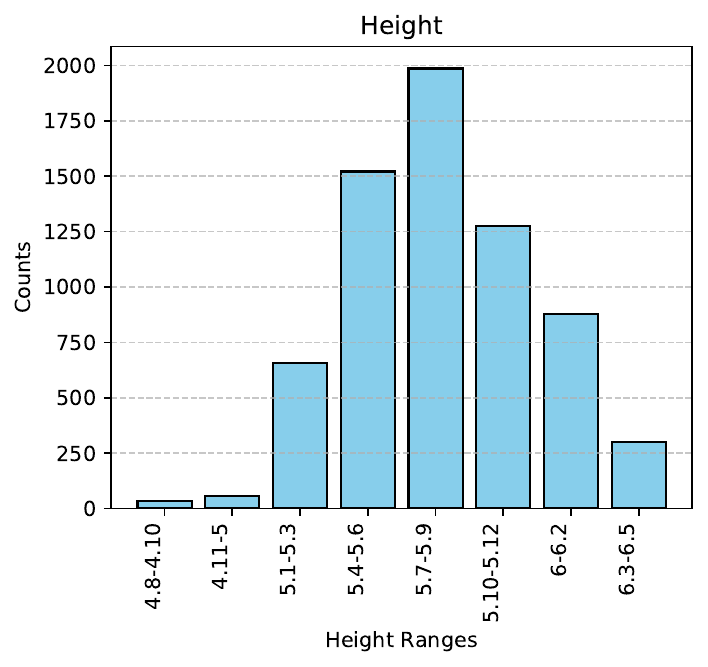}
        \caption{Dataset (Height)}
        \label{fig:4}
    \end{minipage}
    \hfill 
    \begin{minipage}[b]{5.2cm} 
        \centering
        \includegraphics[width=\linewidth, height=5.2cm]{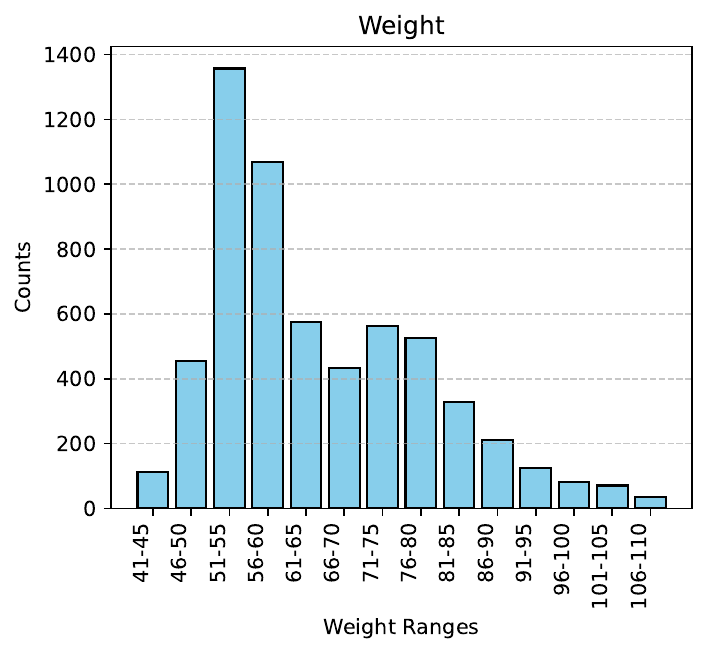}
        \caption{Dataset (Weight)}
        \label{fig:5}
    \end{minipage}
\end{figure*}

\section{Methodology}
\begin{figure*}
 \centering
\centerline{\includegraphics[width=400pt,height=190pt]{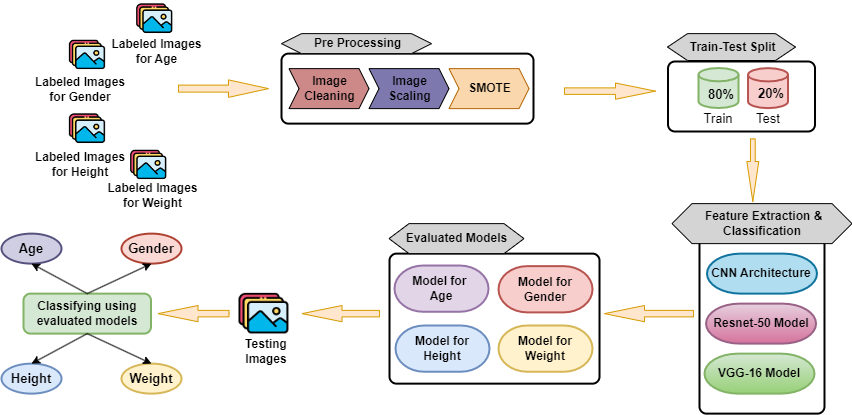}}
\caption{
 Proposed methodology for age, gender, height and weight estimation using different deep learning models }
\label{Figure:1}
\end{figure*}

We proposed a process in this section for estimating age, gender, height and weight. The proposed method for classification is schematically represented in
Fig.~\ref{Figure:1}.

\subsection{Input data}
This system utilized the dataset's images as input where each containing labeled information for height, weight, age, and gender.

\subsection{Preprocessing}
We performed image cleaning, scaling, Synthetic Minority Over-sampling Technique (SMOTE) \cite{b17} and shuffling as part of the preprocessing procedures. In order to perform image cleaning, we first kept images that had all four attributes information and removed poor quality images. Following this cleaning process, the dataset was reduced from 9500 to 7211 data subjects. We resized the images to 224 by 224 pixels after cleaning. Due to the imbalance in the dataset, we applied SMOTE on images to balance and enhance classification performance. In \cref{tab:table3}, we observed that the application of SMOTE resulted in an increase in the number of images for four attributes. Also we performed shuffling to ensure unbiasedness in training.

\subsection{Feature extraction and classification}
In our approach, we employed CNN architecture \cite{b18}, pre-trained ResNet-50
\cite{b19} and VGG-16 \cite{b20} as feature extractors for four attributes (Age, Gender,Height and Weight) and also as classifiers for classification.
\subsection{Evaluated Models}
After completing the classification task, we assessed the three models' performance for each of the four attributes. The best-performing model was chosen for each attribute. The selected model was named the age model for age classification, the gender model for gender classification, the height model for height classification, and the weight model for weight classification.

\section{Evaluation Matrics}
We used a number of metrics such as accuracy, recall, precision and F1-score to evaluate the suggested system's classification performance. True negative (tn), true positive (tp), false negative (fn) and false positive (fp) were among the parameters used in the evaluation \cite{b21}.
\subsection{\textbf{Accuracy}}
Accuracy is a measure that reflects the proportion of correctly classified instances \cite{b22}.
\begin{equation}
\text { Accuracy }=\frac{(tn+tp)}{(tn+tp+fn+fp)}
\end{equation} 
\subsection{\textbf{Precision}}
It is defined as the ratio of precisely anticipated occurrences to the total number of actual occurrences \cite{b23}.
\begin{equation}
\text { Precision }=\frac{tp}{(fp+tp)}
\end{equation} 
\subsection{\textbf{Recall}}
It includes figuring out the ratio of accurately anticipated cases to all expected occurrences \cite{b24}.
\begin{equation}
\text { Recall }=\frac{tp}{(tp+fn)}
\end{equation} 
\subsection{\textbf{F1-score}}
It is a single metric—mathematically represented as the harmonic mean of recall and precision—that takes accuracy and recall into account \cite{b25}.
\begin{equation}
\text { F1-Score }=2 \times \frac{( recall  \times precision )}{(  recall + precision )}
\end{equation} 
\newline
\section{Experimental Results}
\subsection{Experimental Setup}
A V100 GPU with a computing capacity of 100 units provided the computational resources for the research which was conducted on the Google Colab Pro platform.
\subsection{Hyper parameter settings}
The dataset was divided into two groups: training (80\%) and testing (20\%). In addition, we trained each model with a batch size of 32 and 20 epochs.

\subsection{Results}
\begin{figure*}[htbp] 
    \centering
    \begin{minipage}[b]{0.48\linewidth} 
        \centering
        \includegraphics[width=\linewidth]{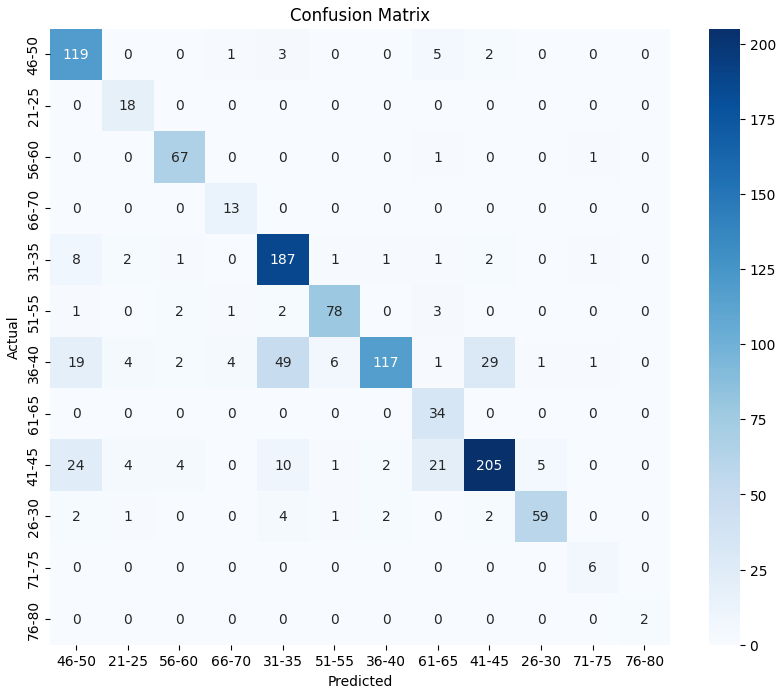}
        \caption{Confusion matrix for Age}
        \label{fig:7}
    \end{minipage}
    \hfill 
    \begin{minipage}[b]{0.48\linewidth} 
        \centering
        \includegraphics[width=\linewidth]{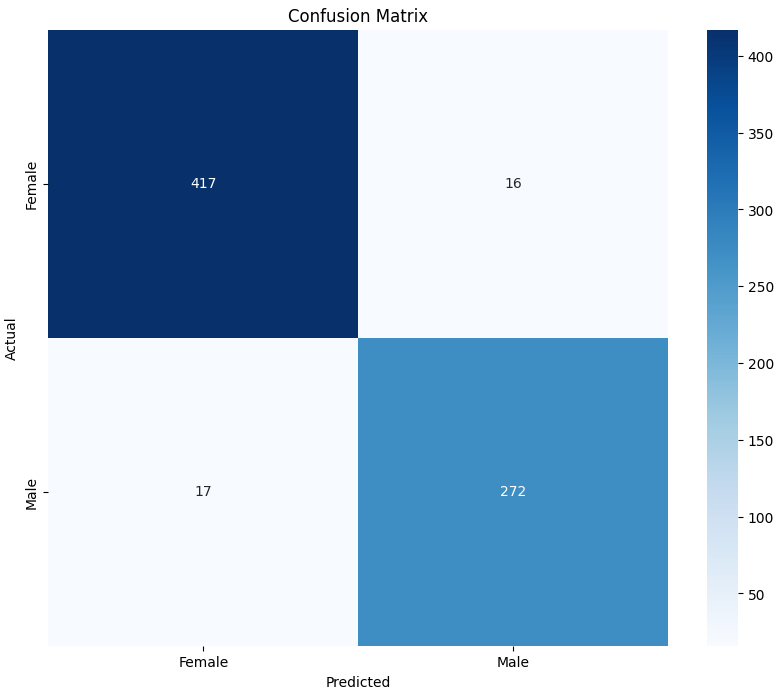}
        \caption{Confusion matrix for Gender}
        \label{fig:8}
    \end{minipage}
\end{figure*}

\begin{figure*}[htbp] 
    \centering
    \begin{minipage}[b]{0.48\linewidth} 
        \centering
        \includegraphics[width=\linewidth]{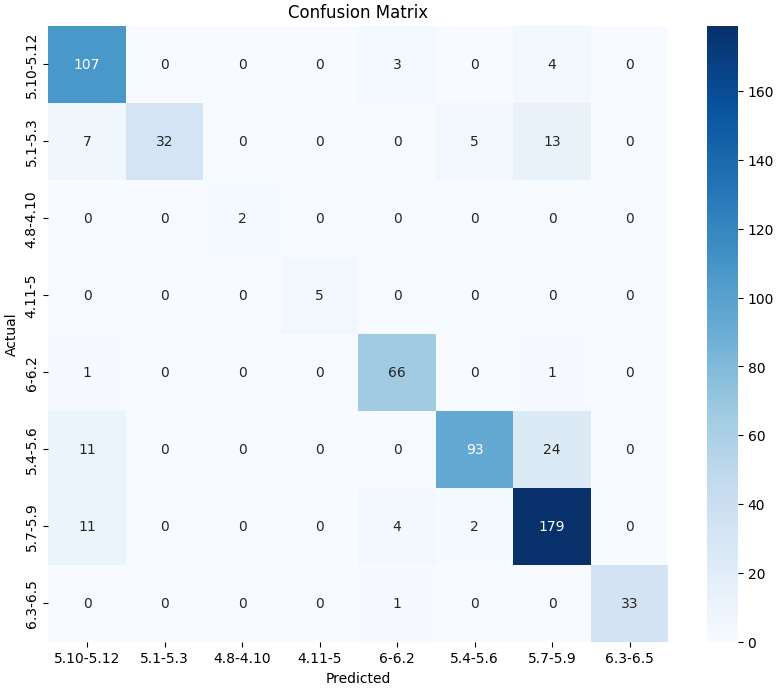}
        \caption{Confusion matrix for Height}
        \label{fig:9}
    \end{minipage}
    \hfill 
    \begin{minipage}[b]{0.48\linewidth} 
        \centering
        \includegraphics[width=\linewidth]{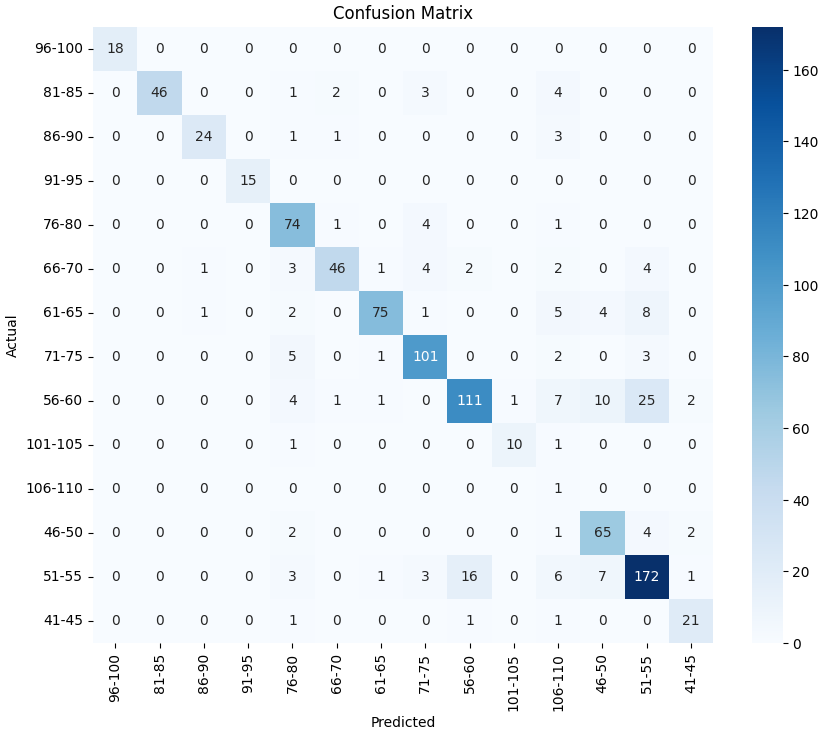}
        \caption{Confusion matrix for Weight}
        \label{fig:10}
    \end{minipage}
\end{figure*}

We used deep learning models namely CNN, ResNet-50, and VGG-16 as classifiers to classify the four attributes: age, gender, height and weight. Following the classification process, we present the confusion matrix in Fig.~\ref{fig:7} -- Fig.~\ref{fig:10} for the most effective model among the three along with an evaluation table detailing the performance for each attribute on test data.

\begin{table}
\centering
\caption{Result Analysis}
\label{tab:table4}
\begin{tabular}{ |c|c|c|c|c|c| } 
\hline
\textbf{Attributes} & \textbf{Models} & \textbf{Accuracy} & \textbf{Precision} & \textbf{Recall} & \textbf{F1-score}\\
\hline
\multirow{3}{*}{\textbf{Age}} & CNN & 26.77\% & 8.66 & 8.58 & 5.64 \\ 
& \textbf{Resnet-50} & \textbf{79.18\%} & \textbf{78.39} & \textbf{89.80} & \textbf{81.67} \\ 
& VGG-16 & 39.72\% & 8.64 & 8.60 & 7.30 \\ 
\hline
\multirow{3}{*}{\textbf{Gender}} & CNN & 80.05\% & 80.41 & 77.56 & 78.36 \\ 
& \textbf{Resnet-50} & \textbf{95.43\%}& \textbf{95.26} & \textbf{95.21} & \textbf{95.24} \\ 
& VGG-16 & 93.35\% & 93.44 & 92.67 & 93.02 \\ 
\hline
\multirow{3}{*}{\textbf{Height}} & CNN & 56.29\% & 78.13 & 64.20 & 66.94 \\ 
& \textbf{Resnet-50} & \textbf{85.60\%} & \textbf{92.66} & \textbf{88.51} & \textbf{89.51} \\ 
& VGG-16 & 46.36\% & 63.62 & 61.71 & 57.25 \\ 
\hline
\multirow{3}{*}{\textbf{Weight}} & CNN & 25.66\% & 16.41 & 11.52 & 8.90 \\ 
& \textbf{Resnet-50} & \textbf{81.91\%} & \textbf{82.57} & \textbf{86.30} & \textbf{80.78} \\ 
& VGG-16 & 42.59\% & 59.74 & 52.08 & 53.14 \\ 
\hline
\end{tabular}
\end{table}


Analysing the \cref{tab:table4}, it can be concluded that in the assessment of classification models across four attributes - age, gender, height, and weight - CNN displayed poor performance, particularly struggling with age, gender, and weight classifications. Similarly, the VGG16 model showed inadequacy, especially in accurately predicting heights. In contrast, ResNet50 consistently outperformed both CNN and VGG16 across all attributes. The evaluation metrics further confirmed ResNet50's superiority, with an accuracy of 79.18\% for age, 95.43\% for gender, 85.60\% for height, and 81.91\% for weight. The precision, recall, and F1-score also demonstrated ResNet50's dominance, boasting values of 78.39, 95.26, 92.66, and 82.57 for precision; 89.80, 95.21, 88.51, and 86.30 for recall; and 81.67, 95.24, 89.51, and 80.78 for F1-score, respectively. These results demonstrate ResNet50's improved performance over CNN and VGG16 and affirm it as the most dependable model for classification across all attributes. 

\section{Future work}
There exists potential for performing adjustments to achieve a better balance in the ‘Celeb-FBI’ dataset. Utilizing this dataset, a model that can predict BMI and BMR directly from full body images can be developed and revolutionize health assessments. The proposed dataset containing annotated human images is expected to become a valuable resource in several medical fields, providing transformative insights for diagnosis and health decision-making.
\section{Conclusion}
The ‘Celeb-FBI’ dataset represents a significant advancement in computer vision, especially for biometric recognition from full-body images. This contribution fills a notable gap in the availability of such datasets and represents a substantial milestone in this research field. This dataset serves as a crucial resource for researchers working in a variety of fields, including security authentication and healthcare applications. The utilization of deep learning models, specifically ResNet-50, has demonstrated promising results in predicting gender, age, height, and weight from these images. While attempts with other models such as CNN and VGG-16 yielded less satisfactory results. Our publicly available dataset sets the stage for further research in this area.
\subsubsection{\ackname} We would like to express our deepest gratitude to our respected supervisor, for his invaluable contributions, insightful inputs, dedication and support throughout the research. We are also thankful to the expert for his assistance in data validation.
\subsubsection{Author's Contributions.}
This work was carried out in close collaboration among all authors; however, the initial idea was conceived by PD, UAR, BKR and MAR. PD and UAR developed the methods and conducted the experiments. BKR and AHTA contributed to the data analysis. All authors participated in the data collection and manuscript writing process. MAR supervised the overall study and reviewed the manuscript.

\end{document}